\documentclass[conference]{IEEEtran}

\usepackage{algorithm}
\usepackage{algpseudocode}
\usepackage{float}
\usepackage{flushend}
\usepackage{graphicx}
\usepackage{lettrine}
\usepackage{multirow}
\usepackage{subfigure}

\begin{document}

\title{Evolutionary Artificial Neural Network Based on Chemical Reaction Optimization}

\author{James J.Q. Yu\\
       Department of Electrical and\\
       Electronic Engineering\\
       The University of Hong Kong\\
       Email: jamesyjq@hku.hk\\
\and Albert Y.S. Lam, \textit{Member, IEEE}\\
       Department of Electrical Engineering\\
       and Computer Sciences\\
       University of California, Berkeley\\
       Email: ayslam@eecs.berkeley.edu\\
\and Victor O.K. Li, \textit{Fellow, IEEE}\\
       Department of Electrical and\\
       Electronic Engineering\\
       The University of Hong Kong\\
       Email: vli@eee.hku.hk\\
}

\maketitle
\pagestyle{empty}

\begin{abstract}
Evolutionary algorithms (EAs) are very popular tools to design and evolve artificial neural networks (ANNs), especially to train them. These methods have advantages over the conventional backpropagation (BP) method because of their low computational requirement when searching in a large solution space. In this paper, we employ Chemical Reaction Optimization (CRO), a newly developed global optimization method, to replace BP in training neural networks. CRO is a population-based metaheuristics mimicking the transition of molecules and their interactions in a chemical reaction. Simulation results show that CRO outperforms many EA strategies commonly used to train neural networks.
\end{abstract}


\section{Introduction}

\lettrine[lines=2]{A}{rtificial} neural networks (ANNs) are complex networks imitating the way human cerebral neurons process information to realize parallel information transformation and processing. ANNs have been widely employed to solve real life problems related to classification, function approximation, data processing, and robotics. The training algorithm used to determine various parameters of ANN is one of the key factors that influence the performance of ANNs. Among all training algorithms, backpropagation (BP) has been widely used, but it suffers from the problem that it is easy to get stuck in the local optima, and its low convergence speed \cite{CKGoh2008}\cite{XinYao1997}. With the advancement of evolutionary algorithms (EAs), they are employed to train ANNs. Moreover, an EA can simultaneously optimize the weights of an ANN and it can also evolve the structure of the network so as to achieve desirable performance \cite{KStanley2002}.

With different levels of EA involvement, EA-based ANNs can be classified into two major types: ``noninvasive" and ``invasive". The former refers to those methods using EA to evolve network structure in conjunction with BP for weight adaptation. The latter includes those using EA for both evolving network structure and weight adaptation \cite{PPPalmes2005}. Due to the tradition of employing BP for network weight training, ``noninvasive" methods have been widely studied and many algorithms have been developed with outstanding performance \cite{FLeung2003}\cite{PACastillo1999}. Since they rely heavily on BP, they still suffer from the problems of getting stuck in local optima and low convergence speed \cite{MGori1992}.

The ``invasive" methods, however, merely depend on EA for evolving the network. Thus the computation speed is higher than ``noninvasive" methods since the ``invasive" methods can avoid BP fitness evaluation with direct representation of networks. In this paper, we propose a new algorithm based on Chemical Reaction Optimization (CRO) \cite{AYSLam2010} to evolve the network structure and to tune the weights of networks.

CRO is a novel chemical reaction-inspired general purpose optimization algorithm. It is a variable population-based metaheuristics, mimicking the transition of molecules and intermolecular interactions in a chemical reaction. The transitions and interactions tend to direct molecules toward the lowest potential energy states on the potential energy surface (PES). Thus CRO uses the idea of mimicking the objective function landscape with PES and molecules can explore the solution space to find the global optimum due to this tendency. CRO has been proved to be effective in solving many practical problems \cite{AYSLam2010}\cite{AYSLam2010b}\cite{JinXu2010} and simulation results show that ANNs trained by CRO outperform other EAs in many classification problems.

The rest of the paper is organized as follows. In Section II, we briefly present the related work of using EA to train ANNs. In Section III, the problem formulation is presented. The detailed framework and the algorithm based on CRO is given in Section IV. Section V presents the simulation results comparing CRO-based ANNs (CROANN) with other ANNs. Finally we conclude the paper in Section VI.

\section{Related Work}

Using EA to train ANNs has become an active research topic. Many EAs, e.g. genetic algorithm (GA) \cite{JTTsai2006}, simulated annealing (SA) \cite{TBLudermir2006}, and particle swarm optimization (PSO) \cite{JianboYu2007} have been used. Yet relatively few ``invasive" methods have been studied to achieve the best performance of EA-based neural networks. Sexton \textit{et al.} used Tabu Search (TS) for neural network training \cite{RSSexton1998}, where TS was used to train a fixed neural network with six hidden layer neurons. The TS solution is given in the form of vectors representing all the weights of the network. The testing data set was a collection of randomly generated two-dimensional points $(x,y)$ where $x\in \lbrack -100,100]$ and $y\in \lbrack -10,10]$. The output data set was generated by simple mathematical functions. The result demonstrated that TS-based networks could outperform conventional BP-derived networks. SA and GA were also implemented for the same data set \cite{RSSexton1999}.

Angeline \textit{et al.} proposed GeNeralized Acquisition of Recurrent Links (GNARL) using hybrid-GA to train ANNs \cite{PJAngeline1994}. Instead of using symmetric topology, GNARL employs sparse connections of neural networks to represent the network structures. GNARL uses a mutation operation to evolve the structure and to tune the weights of networks. GNARL reserves the top 50\% individuals in each generation, according to the user-defined fitness function, and performs reproduction by two types of mutation methods: parametric mutation and structural mutation. The former mutation method changes the network by perturbing the weights with Gaussian noise controlled by an annealing temperature \cite {SKirk1984}, while the latter mutation method involves the addition or deletion of hidden layer nodes or links.

A Constructive algorithm for training Cooperative Neural Network Ensembles (CNNE), proposed by M. Islam  \textit{et al.} \cite{MIslam2003}, uses a constructive algorithm to evolve neural networks. CNNE relies on the contribution of individuals in the population and uses incremental learning to maintain the diversity among individuals in an ensemble. Incremental learning based on negative correlation could effectively reduce the redundancy generated by individuals searching the same solution space and thus different individuals could learn different aspect of the training data, which could result in a final solution of the ensemble. CNNE is a ``noninvasive" method which relies on proper implementation of BP. Though CNNE minimizes optimization problems by utilization of ensembles, it suffers from the ``structural climbing problem" \cite{PJAngeline1994}.

S. He  \textit{et al.} proposed a Group Search Optimizer-based ANN (GSOANN) \cite{SHe2009}, which uses Group Search Optimizer, a population-based optimization algorithm inspired by animal social foraging behavior, to train the networks with least-squared error function as the fitness function.

Paulito P. Palmes \textit{et al.} proposed mutation-based genetic neural network (MGNN) employing a specially designed mutation strategy to perturb the chromosomes representing neural networks \cite{PPPalmes2005}. MGNN is very similar to GNARL except that it implements selection, encoding, mutation, fitness function, and stopping criteria differently. MGNN's encoding scheme contributes to a flexible formulation of fitness function and mutation strategy of local adaptation of evolutionary programming, and it implements a stopping criteria using ``sliding window" to track the state of overfitness.

\section{Problem Formulation}

In this paper, we consider the problem of single-hidden-layer feedforward neural network (SLFN\footnote{SLFN is the original abbreviation used in \cite{CKGoh2008} to refer to a ``single-hidden-layer feedforward neural network".}) design for data classification. We use a topological structure, activation functions of nodes, and connection weights to describe an SLFN. We use $l\in[0,1,2]$ to distinguish the different layers, where Layer 0 to 2 are the input layer, the hidden layer, and the output layer, respectively. $n_{l}$ stands for the number of neurons in Layer $l$. $w_{k,pq}$ represents the weight of the connection between the $p^{th}$ neuron in Layer $k-1$ to the $q^{th}$ neuron in Layer $k$. $b_{k,p}$ stands for the bias of the $p^{th}$ neurons in Layer $k$. Fig. 1 depicts an example of SLFN. In the problem dataset $S$ with $|S|$ samples, the $i^{th}$ sample $i$ is composed of a pattern $S_{i}$ and the corresponding desired output $R_{i}$, where $i=1,2,...,|S|$. Thus we use $s_{i,m}$ to describe the $m^{th}$ element of the $i^{th}$ pattern, and $r_{i,q}$ to denote the $q^{th}$ element of the $i^{th}$ desired outputs in the dataset. With a given SLFN and a pattern $S_{i}$, the computed result $c_{i,q}$ can be obtained from the following formula
\begin{equation}
  c_{i,q}=f_{2}(\sum_{p=1}^{n_{1}}(w_{2,pq}\times f_{1}(\sum_{m=1}^{n_{0}}(w_{1,mp}\times s_{i,m})+b_{1,p}))+b_{2,q}) \label{classR}
\end{equation}
where $f_{1}$ and $f_{2}$ are the activation functions for hidden layer neurons and output neurons, respectively. For a trained SLFN, we can make use of the difference between $R_{i}$ and $C_{i}$ to evaluate the performance of the network.

\begin{figure}[t]
  \centering
  \includegraphics{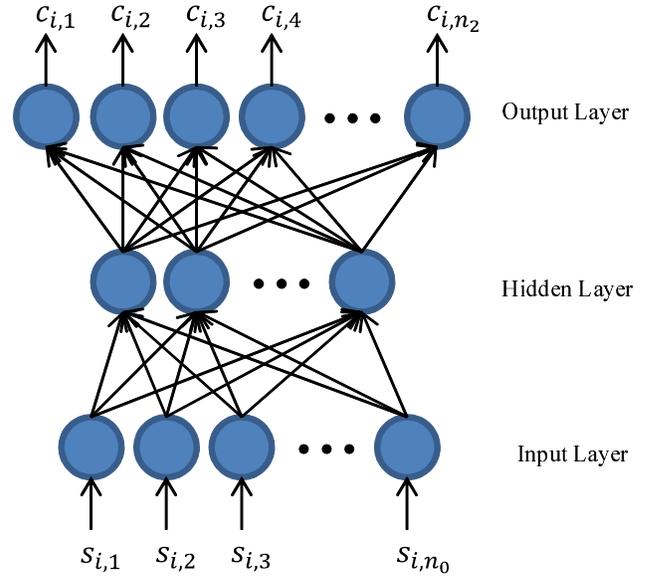}
  \caption{Single-hidden-layer Feedforward Neural Network}
\end{figure}

The primary function to evaluate the performance of a conventional BP network is the mean squared error (MSE) between $R_{i}$ and $C_{i}$. A small MSE means that the performance is good and the network is well-trained. However, in order to concentrate more on the accuracy of the classification result, here we adopt a new fitness function $f_{fitness}$ (\ref{Ffit}) consisting of the normalized mean squared error (NMSE) $f_{NMSE}$ (\ref{Fnmse}) and the classification percentage error $f_{percent}$ (\ref{Fpercent}). Their formulas are given as follows:
\begin{equation}
  f_{fitness}=\alpha \times f_{NMSE}+\beta \times f_{percent}  \label{Ffit}
\end{equation}
\begin{equation}
  f_{NMSE}=\frac{100}{n_{2}S}\times \sum_{i=1}^{|S|}\sum_{q=1}^{n_{2}}\left(r_{i,q}-c_{i,q}\right) ^{2}  \label{Fnmse}
\end{equation}
\begin{equation}
  f_{percent}=100\times \left( 1-\frac{correct}{total}\right)  \label{Fpercent}
\end{equation}
In (\ref{Ffit}), $\alpha $ and $\beta $ are user-defined parameters to balance the weighting on (\ref{Fnmse}) and (\ref{Fpercent}) in the ultimate fitness function and should be set to a small real value between 0 and 1. For instance, implementations stressing classification correctness can set $\alpha =1$ and $\beta =0.7$. In this paper, we also use this setting in the simulation.

For CROANN, we divide the samples in a dataset into a training set, a validation set, and a testing set. We first determine the best \{$w_{1,mp},w_{2,pq},b_{1,p},b_{2,q}$\} quadlet with the training set according to (\ref{Ffit}). Then we use the validation set to detect and avoid overfitness (See Section IV.D). Since ANNs should have the ability to process unfamiliar data, we use the testing set to evaluate the accuracy of classification and compare the results with other algorithms.

\section{Algorithm Design}

In this section, we first discuss how CRO works. Then we introduce the encoding scheme and operators employed to train the structure and weights of networks for CROANN. Finally the stopping criteria is given.

\subsection{Chemical Reaction Optimization}

CRO is a population-based metaheuristic inspired by chemical reactions, mimicking the process of reactions where molecules collide with the walls of the container and with each other. In the process, molecules attempt to reach the stable state. Imagine we put a certain number of molecules in a closed container. At the start of reaction, molecules with excess energy are unstable. Since there is a natural tendency for a reacting system to stay in a stable state, molecules change their energy state from high to low through a sequence of elementary reactions. When the reaction stops, we can get molecules with the minimum stable state of energy. If we consider different energy states as an energy surface, the transition and interaction of molecules can result in a gradual rolling down process on PES and the lowest point is the minimum stable state of energy. We call the initial molecules ``reactants" and the final ``products".

In CRO, each molecule has a molecular structure, representing a solution of the problem, and two kinds of energy, i.e. potential energy (PE) and kinetic energy (KE). PE stands for the fitness function value and KE describes the tolerance of a molecule to an increase of its energy state. Suppose $\omega $ and $f$ are a molecular structure and the fitness function, respectively, then we compute its PE with $PE_{\omega }=f\left( \omega \right) $. If $\omega ^{\prime }$ is the new structure derived from $\omega $ in an elementary reaction, then $PE_{\omega }+KE_{\omega }>PE_{\omega ^{\prime }}$ has to be satisfied. Otherwise, the reaction is invalid and the new structure should be rejected. In other words, KE represents the ability for a molecule to escape from a local minimum. This rule can also be easily applied to intermolecular elementary reactions and changes may be more vigorous since more KE can be transformed into PE. A central energy buffer is set up for energy conservation and convergence.

We define four types of elementary reactions for CRO, namely, on-wall ineffective collision, decomposition, inter-molecular ineffective collision, and synthesis. These four elementary reactions are defined to cover all possible reactions under the framework of CRO. These four types can be classified into two classes: uni-molecular reactions include the first two types and inter-molecular reactions include the latter two. A uni-molecular reaction can be triggered when a single molecule collides on a wall of the container. An inter-molecular reaction happens when two or more molecules collide with each other (for simplicity, only two molecules are considered in this class of reactions). Interested readers can refer to \cite{AYSLam2010} for the pseudocode of CRO.

\subsection{Encoding}

To accelerate the simulation and to reduce the difficulty in programming, we use two matrices and two vectors to represent different weights and thresholds. This scheme is similar to that described in \cite{PPPalmes2005} with one key difference. In \cite{PPPalmes2005}, there is an extra element in each solution which controls the perturbation strength, but we abandon it since CRO uses one constant parameter to control the Gaussian perturbation. We call the complete collection of these matrices and vectors a ``solution structure". Every molecule has a solution structure representing the network structure and determining the current energy state the molecule is at.

\subsection{Operators}

\subsubsection{Initial Solution Generator{}}

This operator is designed to generate a new structure of networks randomly. Each call will generate a new solution structure. It is achieved by first assigning random numbers to all elements and then scaling them to [-1.0, 1.0] linearly. Its pseudocode is given in Algorithm 1 below:

\begin{algorithm}
  \caption{\sc{InitialGen} ($\omega$)}
  \begin{algorithmic}[1]
    \ForAll{Matrices and vectors $m$ in $\omega$}
      \ForAll{Elements $e$ in $m$}
        \State Randomly generate a real number $n$
        \State $e=n$
      \EndFor
      \ForAll{Elements $e$ in $m$}
        \State $e=2*(e-min(m))/(max(m)-min(m))-1$
      \EndFor
    \EndFor
  \end{algorithmic}
\end{algorithm}

\subsubsection{Neighbor Generator}

This operator is designed to generate a new solution $\omega ^{^{\prime }}$ from a given solution $\omega $. Its main purpose is to perform a local search for better solutions. It is done by perturbing one random element in the matrices or vectors in $\omega $ with Gaussian perturbation, whose mean is the original number and variation is a user-defined value. Its pseudocode is given in Algorithm 2 where $\rho $ stands for Gaussian perturbation function and $v$ is a user-defined variance.

\begin{algorithm}
  \caption{\sc{Neighbour} ($\omega$)}
  \begin{algorithmic}[1]
    \State Generate a random integer $i$ smaller than the total number of elements in a solution
    \State Find the $i^{th}$ element $e$ in $\omega$
    \State $e=e+\rho(e, v)$
  \end{algorithmic}
\end{algorithm}

\subsubsection{Decomposition}

This operator is used to generate two different solutions $\omega _{1}^{^{\prime }}$ and $\omega _{2}^{^{\prime }}$ based on a given solution $\omega $. This operator can help molecules jump out of local minimums by performing severe perturbation on the solution. It is done by perturbing every element in the matrices and vectors in $\omega $ with Gaussian perturbation probabilistically, say 50\%. If, though unlikely to happen, nothing is changed during the first stage of the perturbation, this solution will be perturbed by the Neighbor Generator function. Its pseudocode is given in Algorithm 3 and the variables are as defined in the previous section.

\begin{algorithm}
  \caption{\sc{Decomposition} ($\omega$)}
  \begin{algorithmic}[1]
    \State $change=false$
    \State Copy $\omega$ to $\omega_{1}^{'}$ and $\omega_{2}^{'}$
    \ForAll{Matrices and vectors $m$ in $\omega_{1}^{'}$ and $\omega_{2}^{'}$}
      \ForAll{Elements $e$ in $m$}
        \State Generate a real $r$ between 0 and 1
        \If{$r>0.5$}
          \State $e=e+\rho(e, v)$
          \State $change=true$
        \EndIf
      \EndFor
      \If{$change\neq true$}
        $neighbour($the original $\omega_{1}^{'}$ or $\omega_{2}^{'})$
      \EndIf
    \EndFor
  \end{algorithmic}
\end{algorithm}

\subsubsection{Synthesis}

This operator is used to generate a new solution $\omega ^{^{\prime }}$ based on two given solutions $\omega _{1}$ and $\omega _{2}$. It is done by randomly choosing elements from both solutions with equal possibilities to form a new solution. Its pseudocode is given in Algorithm 4.

\begin{algorithm}
  \caption{\sc{Synthesis} ($\omega_{1}$, $\omega_{2}$)}
  \begin{algorithmic}[1]
    \ForAll{Matrices and vectors $m$ in $\omega^{'}$}
      \ForAll{Elements $e$ in $m$}
        \State Generate a real $r$ between 0 and 1
        \If{$r>0.5$}
          \State $e=$counterpart in $m_{1}$
        \Else
          \State $e=$counterpart in $m_{2}$
        \EndIf
      \EndFor
    \EndFor
  \end{algorithmic}
\end{algorithm}

\subsection{Stopping Criteria}

We introduce two stopping criteria to CROANN: maximum function evaluations (FE) and overfitness detection. The maximum FE criterion is a hard limit of CROANN and no simulation could evaluate the fitness function more than this threshold. The design of the other stopping criterion, overfitness detection, is based on the observation that good performance with the training samples may not necessarily result in a good performance with the testing samples. Poor overall performance might be obtained due to over-training the system in the training phase. To address this problem, we employ a ``sliding window" to monitor the presence of overfitness in the network. CROANN measures the validation performance of the current best network $ValFitness_{i}$ and compares this performance with the previous best validation performance using $ValBest_{i-1}=min(ValFitness_{j},\forall j<i)$ at the end of $i^{th}$ window. If the previous validation performance is better, then we say this network is ``overfitting" and add 1 to the overfitness counter. When this overfitness counter reaches a user-defined threshold, CROANN will terminate. However, once $ValBest_{i}$ is smaller than $ValFitness_{i-1}$, the overfitness counter is reset to zero. The pseudocode describing the stopping criteria is given in Algorithm 5 below:

\begin{algorithm}
  \caption{\sc{Stopping Criteria} ($ValBest$, $CurrentNet$)}
  \begin{algorithmic}[1]
    \If{Current FE exceeds maximum FE}
      \State CROANN stop
    \EndIf
    \State Calculate the validation performance $ValFitness$
    \If{$ValFitness<ValBest$}
      \State $overFitCount=0$
      \State $ValBest=ValFitness$
      \State Store $CurrentNet$
    \Else
      \State $overFitCount++$
      \If{$overFitCount>overFitThres$}
        \State Restore the saved best network
        \State CROANN stop
      \EndIf
    \EndIf
  \end{algorithmic}
\end{algorithm}

\section{Simulation Results}

\begin{figure*}
  \centering
  \mbox{\subfigure{
  \begin{minipage}[t]{0.33\textwidth}
    \includegraphics[width=1\textwidth]{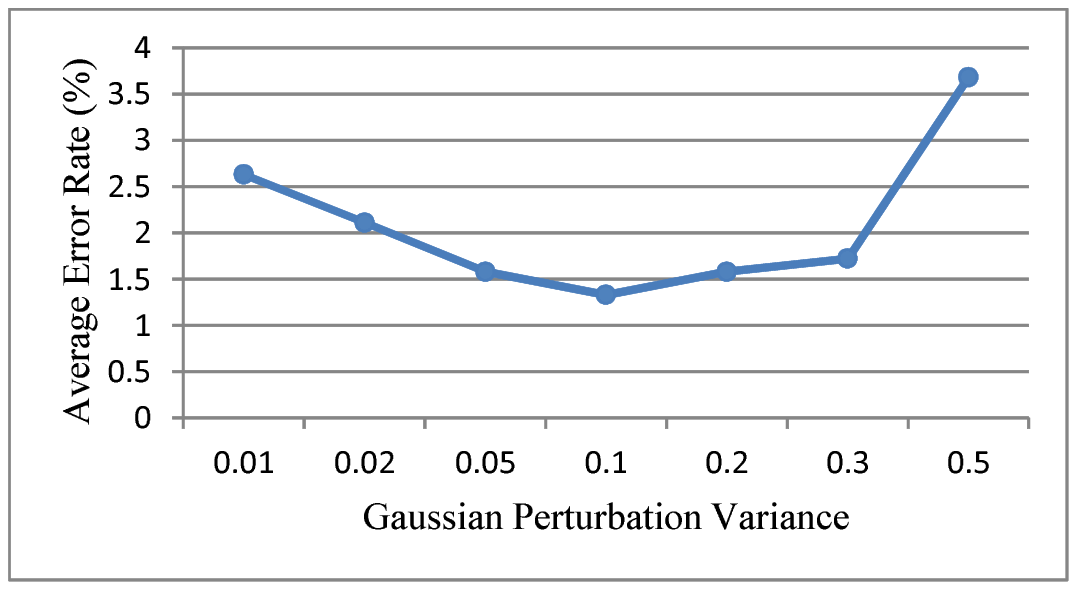} \\\\
    \includegraphics[width=1\textwidth]{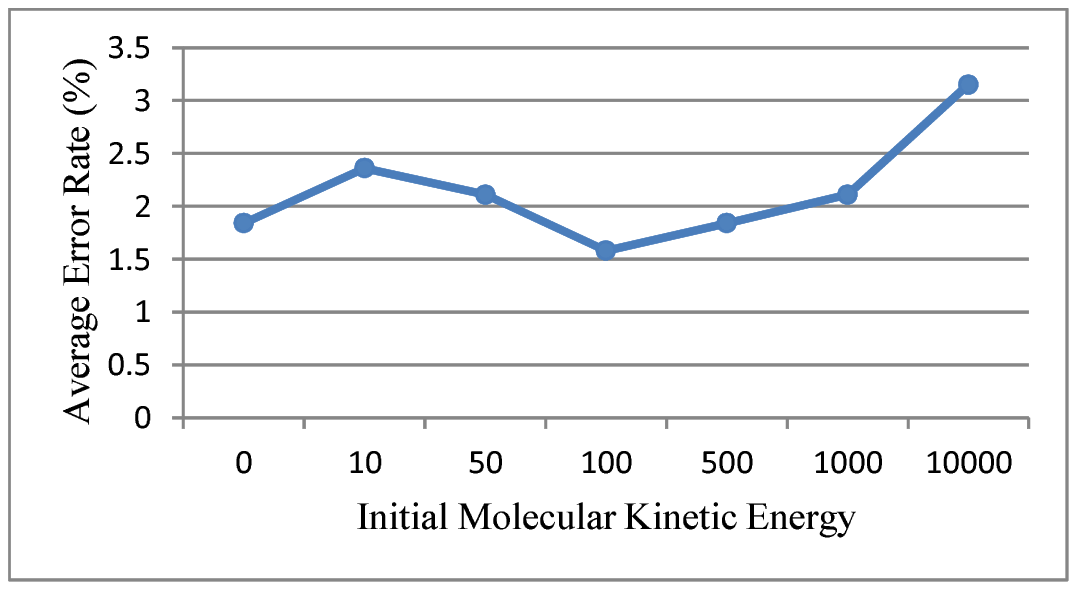} \\\\
    \includegraphics[width=1\textwidth]{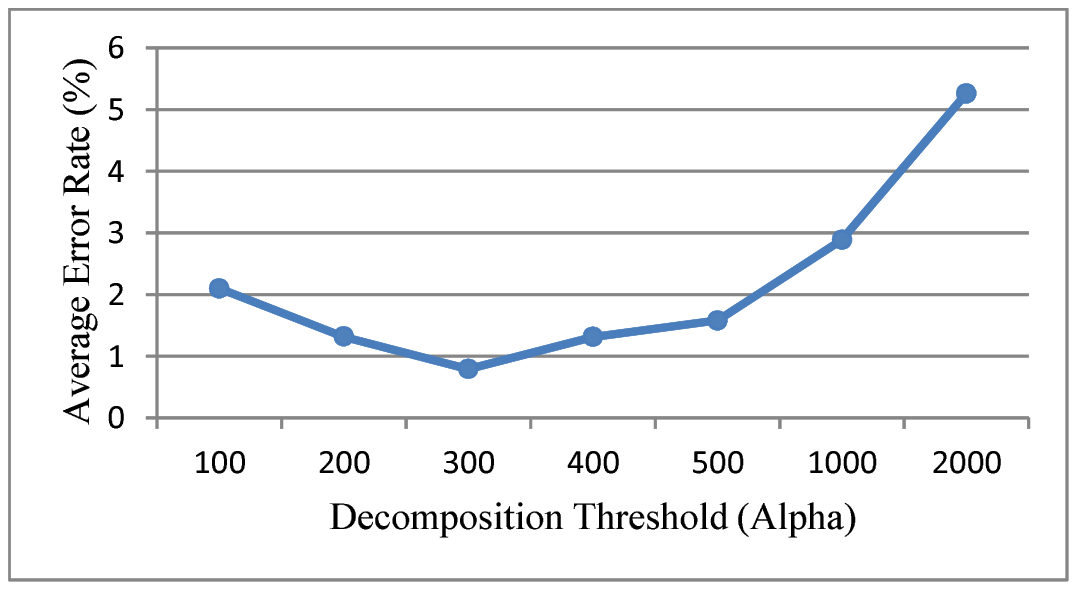}
  \end{minipage}}
  \subfigure{
  \begin{minipage}[t]{0.33\textwidth}
    \includegraphics[width=1\textwidth]{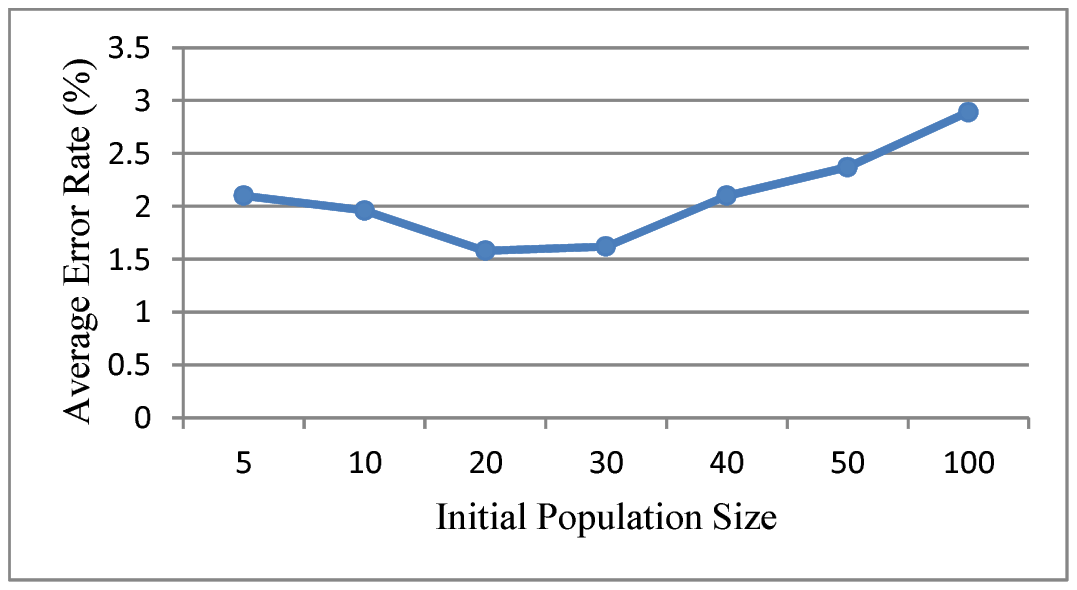} \\\\
    \includegraphics[width=1\textwidth]{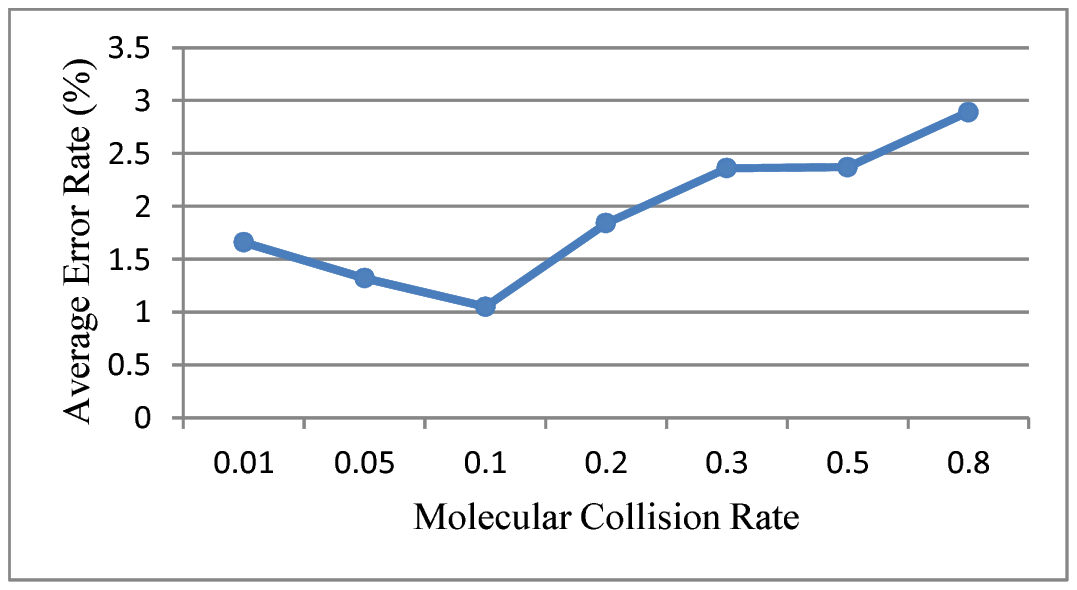} \\\\
    \includegraphics[width=1\textwidth]{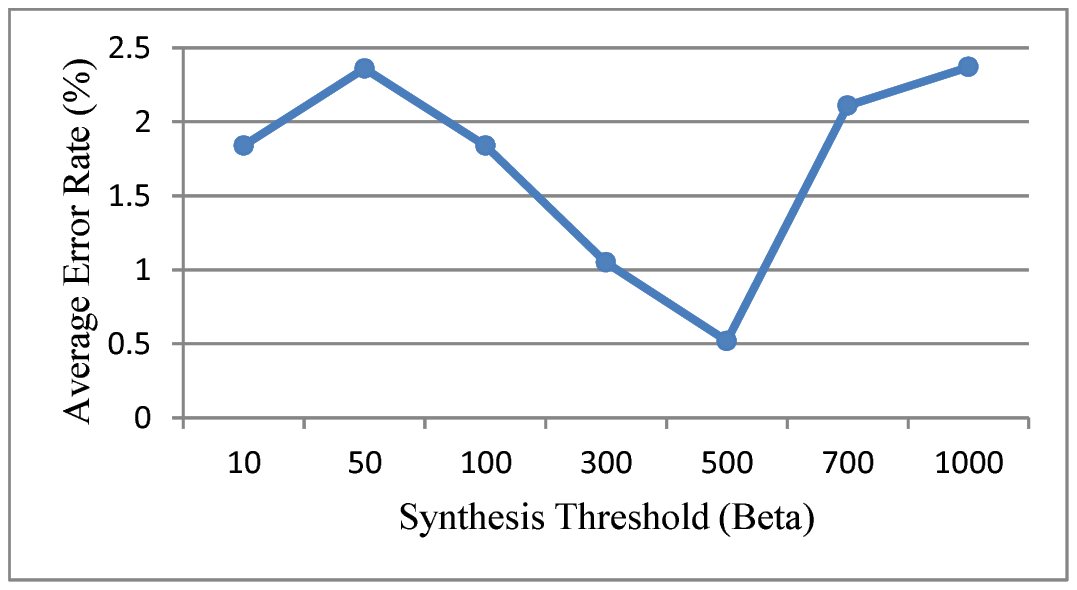}
  \end{minipage}}
  \subfigure{
  \begin{minipage}[t]{0.33\textwidth}
    \includegraphics[width=1\textwidth]{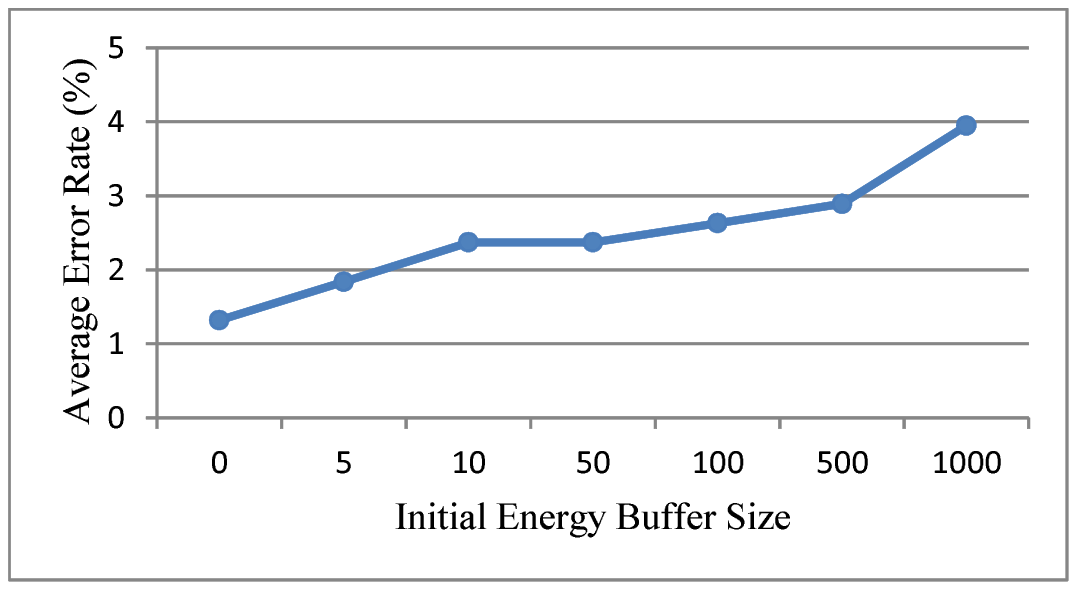} \\\\
    \includegraphics[width=1\textwidth]{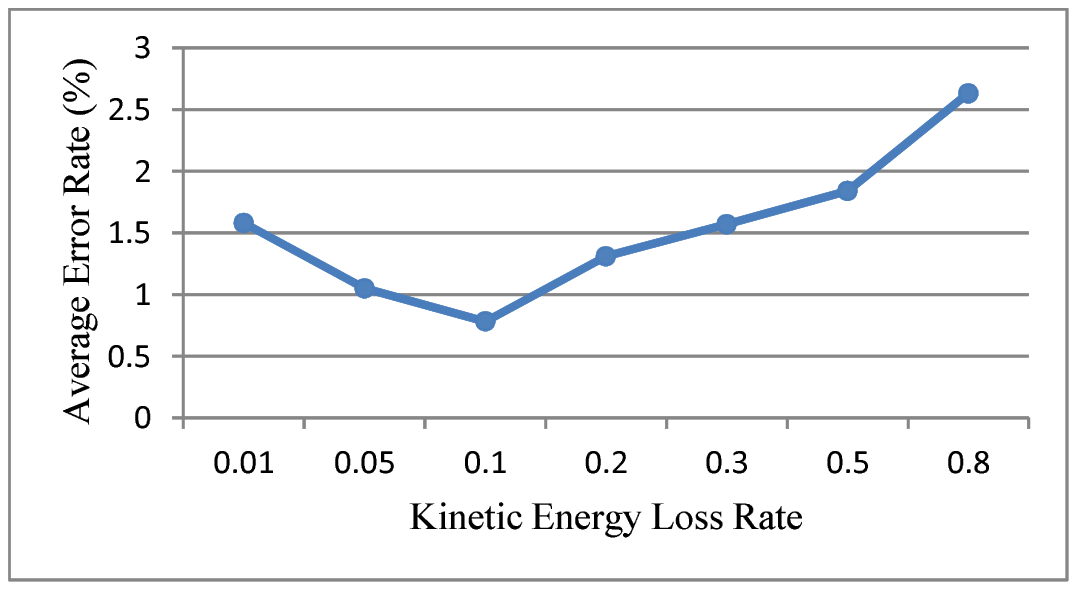}
  \end{minipage}}}
  \caption{Impact of CRO Parameters on Average Error Rate of Classification}
\end{figure*}

In order to evaluate the performance of CROANN, the simulation is implemented with C++ and tested with some famous classification datasets from the UCI repository \cite{FrankA2010}:  \textit{Iris} classification dataset, \textit{Wisconsin breast cancer} classification dataset and \textit{Pima Indians diabetes} dataset. They are all derived from real-world problems. The Iris dataset is a standard benchmark for evaluating the performance of ANNs and has been tested by many neural network algorithms, including some EA-based ANNs algorithm. The latter two datasets are used to test the ability of CROANN to recover from polluted data \cite{XinYao1997}. The datasets are partitioned based on the suggestion given by Prechelt \cite{Prechelt1994}. Each of them is separated into three classes: training class, validation class and testing class with the ratio of 2:1:1 using the simple random sampling method. First, several tests are conducted to evaluate the impact of changes in different CRO parameters, using the \textit{Iris} classification dataset as benchmark. Then CROANN is compared with other EA-based ANN algorithms proposed in the recent literature to evaluate the performance.

\subsection{Analysis of the impact of CROANN parameters}

The ratio of occurrence for different elementary reactions of CRO and the initial size of energy buffer have direct impact on the final performance of the neural network. So it is essential to analyze their impact in order to adjust them for later use. The first experiment includes a set of tests on different CRO parameter values, based on the \textit{Iris} dataset, a benchmark test for machine learning and pattern recognition. When investigating one parameter, other parameters are set constant. Results of each test are generated by averaging the error rate of testing set in 50 trials. The analysis results are shown in Fig. 2.

Results show that a Gaussian perturbation variance which is too large makes CRO scan through the solution space in a relatively large scale but can not explore small regions carefully while a variance which is too small is likely to result in getting stuck in local optima. Similarly, a small population can not let the molecules fully explore the solution space, while a large population will reduce the possibility of reaction occurrence of individual molecules. Initial energy buffer size, initial molecular kinetic energy, and kinetic energy loss rate control the overall energy in the whole system. They also cooperate with molecular collision rate, thresholds for decomposition and synthesis to control the ratio of occurrence for different elementary reactions of CRO to a proper value.

\subsection{Comparing CROANN with other EA-based ANN algorithms}

\begin{table}
  \caption{CROANN Parameters}
  \small
  \begin{center}
  \begin{tabular}{r||l|l|l}
    \hline\hline
    \multirow{2}{*}{Parameter} & \multicolumn{3}{c}{Value} \\ \cline{2-4}
    & Iris & Cancer & Diabetes \\ \hline\hline
    Function Evaluation Limit & 50 000 & 50 000 & 172 800 \\
    Max Window Count & 300 & 300 & 500 \\  \hline
    Gaussian Perturbation Variance & \multicolumn{3}{c}{0.1} \\
    Initial Population Size & \multicolumn{3}{c}{20} \\
    Initial Energy Buffer Size & \multicolumn{3}{c}{0.0} \\
    Initial Molecular Kinetic Energy & \multicolumn{3}{c}{100.0} \\
    Molecular Collision Rate & \multicolumn{3}{c}{0.1} \\
    Kinetic Energy Loss Rate & \multicolumn{3}{c}{0.1} \\
    Decomposition Threshold & \multicolumn{3}{c}{300} \\
    Synthesis Threshold & \multicolumn{3}{c}{500} \\
    Number of Trials & \multicolumn{3}{c}{50} \\
    Window Size & \multicolumn{3}{c}{100} \\ \hline\hline
  \end{tabular}%
  \end{center}
\end{table}

\begin{table*}
  \caption{Error Rate (\%) of CROANN and Other ANNs of the Iris Dataset}
  \small
  \begin{center}
  \begin{tabular}{r||llll||llll||llll}
    \hline\hline
    \multirow{2}{*}{Algorithm} & \multicolumn{4}{c||}{Training Set} & \multicolumn{4}{c||}{Validation Set} & \multicolumn{4}{c}{Testing Set} \\ \cline{2-13}
    & Mean & Std. & Min & Max & Mean & Std. & Min & Max & Mean & Std. & Min & Max \\ \hline\hline
    CROANN & 2.00 & 3.68 & 0.00 & 5.33 & 4.32 & 2.16 & 2.70 & 8.10 & \textbf{1.31} & 1.77 & 0.00 & 7.89 \\
    SGAANN & 16.24 & 5.92 & 7.21 & 30.23 & - & - & - & - & 14.20 & 8.82 & 0.00 & 36.00 \\
    EPANN & 18.54 & 6.47 & 7.69 & 29.77 & - & - & - & - & 12.56 & 8.42 & 0.00 & 32.00 \\
    ESANN & 14.47 & 5.25 & 6.97 & 27.43 & - & - & - & - & 7.08 & 6.40 & 0.00 & 26.00 \\
    PSOANN & 13.27 & 5.39 & 7.38 & 25.84 & - & - & - & - & 10.38 & 9.36 & 0.00 & 32.00 \\
    GSOANN & 12.03 & 1.60 & 8.63 & 15.36 & - & - & - & - & 3.52 & 2.27 & 0.00 & 8.00 \\
    MGNN & - & - & - & - & - & - & - & - & 4.68 & - & - & - \\ \hline\hline
  \end{tabular}
  \end{center}
\end{table*}
\begin{table*}
  \caption{Comparison between CROANN and Other Machine Learning Algorithms of the Iris Dataset}
  \small
  \begin{center}
  \begin{tabular}{r|lllll}
    \hline
    Algorithm & CROANN & GANet-best \cite{ECantuPaz2005} & SVM-best \cite{TVGestel2004} & CCSS \cite{SDzeroski2004} &  \\ \hline
    Error Rate & \textbf{1.31} & 6.40 & 1.40 & 4.40 &  \\ \hline
  \end{tabular}
  \end{center}
\end{table*}
\begin{table*}
  \caption{Error Rate (\%) of CROANN and Other ANNs of the Wisconsin Breast-Cancer Dataset}
  \small
  \begin{center}
  \begin{tabular}{r||llll||llll||llll}
    \hline\hline
    \multirow{2}{*}{Algorithm} & \multicolumn{4}{c||}{Training Set} & \multicolumn{4}{c||}{Validation Set} & \multicolumn{4}{c}{Testing Set} \\ \cline{2-13}
    & Mean & Std. & Min & Max & Mean & Std. & Min & Max & Mean & Std. & Min & Max \\ \hline\hline
    CROANN & 3.89 & 0.72 & 3.21 & 5.61 & 3.54 & 0.42 & 2.86 & 4.00 & 1.06 & 0.67 & 0.00 & 2.29 \\
    SGAANN & 3.88 & 0.63 & 3.04 & 5.63 & 3.86 & 1.14 & 2.59 & 7.82 & 1.50 & 0.72 & 0.00 & 2.85 \\
    EPANN & 3.58 & 0.63 & 3.03 & 6.18 & 3.30 & 1.45 & 1.85 & 8.99 & 1.54 & 1.16 & 0.57 & 6.29 \\
    ESANN & 2.98 & 0.11 & 2.73 & 3.16 & 2.70 & 0.39 & 2.14 & 3.52 & 0.95 & 0.66 & 0.00 & 2.86 \\
    PSOANN & 3.26 & 0.24 & 2.92 & 3.80 & 2.37 & 0.43 & 1.37 & 3.35 & 1.24 & 2.02 & 0.00 & 11.43\\
    GSOANN & 3.35 & 0.09 & 3.26 & 3.56 & 2.17 & 0.21 & 1.93 & 2.89 & \textbf{0.65} & 1.42 & 0.00 & 1.14\\
    MGNN & - & - & - & - & - & - & - & - & 3.05 & - & - & -\\ \hline\hline
  \end{tabular}
  \end{center}
\end{table*}
\begin{table*}
  \caption{Comparison between CROANN and Other Machine Learning Algorithms of the Wisconsin Breast Cancer Dataset}
  \small
  \begin{center}
  \begin{tabular}{r|llllllll}
    \hline
    Algorithm & CROANN & GANet-best \cite{ECantuPaz2005} & SVM-best \cite{TVGestel2004} & CCSS \cite{SDzeroski2004} & COOP \cite{NGPedrajas2005} & CNNE \cite{MIslam2003} & EPNet \cite{XinYao1997} & EDTs \cite{TGDirt2000} \\ \hline
    Error Rate & \textbf{1.06} & \textbf{1.06} & 3.10 & 2.72 & 1.23 & 1.20 & 1.38 & 2.63 \\ \hline
  \end{tabular}%
  \end{center}
\end{table*}

For comparison, we employ six EA-based training algorithms, namely simple genetic algorithm (SGA) ANNs \cite{JH1975}, evolutionary programming (EP) ANNs \cite{LJFogel1965}\cite{DBFogel1995}, evolutionary strategies (ES) ANNs \cite{HPSchwefel1995}, particle swarm optimizer (PSO) ANN \cite{JKennedy1995}, mutation-based neural networks (MGNN) \cite{PPPalmes2005}, and group search optimizer (GSO) ANNs \cite{SHe2009}. We also compare CROANN with some other sophisticated or hybrid machine learning algorithms further to check whether CROANN can be competitive with them. Since there is no agreement on the maximum number of FEs in the previous literature, the maximum number of FEs for the first two datasets is set to 50 000 according to the average maximum FEs given in \cite{PPPalmes2005}, while for the third dataset, it is set to 172 800 according to \cite{SHe2009}. Other CRO parameters are determined based on the analysis in the previous section. Table I gives the collection of CROANN parameters in the simulation.

\subsubsection{Iris Dataset}

The Iris dataset is the most widely-used benchmark for machine learning and pattern recognition. The whole dataset can be divided into three different classes of iris species: \textit{Setosa}, \textit{Versicolour} and \textit{Verginica}. The species of iris can be determined by four attributes of the plants: \textit{sepal length}, \textit{sepal width}, \textit{petal length} and \textit{petal width}. The dataset is divided into three parts: 75 training samples, 37 validation samples and 38 testing samples.

Results generated by CROANN, averaged over 50 trials, and those of six other ANNs are displayed in Table II. It is easy to see that CROANN outperforms all other EA-based ANNs dramatically, in both training error and testing error. The comparison with recent machine learning algorithms in the literature is listed in Table III. CROANN also generates the best result among these algorithms.

\subsubsection{Wisconsin Breast Cancer Dataset}

\begin{table*}
  \caption{Error Rate (\%) of CROANN and Other ANNs of the Pima Indian Diabetes Dataset}
  \small
  \begin{center}
  \begin{tabular}{r||llll||llll||llll}
    \hline\hline
    \multirow{2}{*}{Algorithm} & \multicolumn{4}{c||}{Training Set} & \multicolumn{4}{c||}{Validation Set} & \multicolumn{4}{c}{Testing Set} \\ \cline{2-13}
    & Mean & Std. & Min & Max & Mean & Std. & Min & Max & Mean & Std. & Min & Max \\ \hline\hline
    CROANN & 16.55 & 2.73 & 15.89 & 18.23 & 16.04 & 3.01 & 14.58 & 17.71 & \textbf{19.67} & 5.38 & 17.19 & 23.44 \\
    SGAANN & 17.73 & 0.96 & 16.05 & 20.67 & 16.48 & 1.23 & 14.61 & 19.44 & 24.46 & 3.75 & 20.31 & 35.94 \\
    EPANN & 18.38 & 1.56 & 16.28 & 21.34 & 17.18 & 1.87 & 14.75 & 20.55 & 25.75 & 4.89 & 18.23 & 36.46 \\
    ESANN & 15.85 & 0.28 & 15.32 & 16.37 & 14.26 & 0.35 & 13.34 & 16.51 & 20.93 & 1.76 & 17.19 & 25.52 \\
    PSOANN & 16.25 & 0.19 & 15.76 & 16.77 & 14.74 & 0.47 & 14.20 & 16.51 & 20.99 & 1.47 & 18.23 & 23.96 \\
    GSOANN & 16.43 & 0.21 & 15.97 & 16.80 & 14.82 & 0.21 & 14.37 & 15.21 & 19.79 & 0.96 & 17.19 & 21.88 \\ \hline\hline
  \end{tabular}%
  \end{center}
\end{table*}
\begin{table*}
  \caption{Comparison between CROANN and Other Machine Learning Algorithms of the Pima Indian Diabetes Dataset}
  \small
  \begin{center}
  \begin{tabular}{r|llllllll}
    \hline
    Algorithm & CROANN & GANet-best \cite{ECantuPaz2005} & SVM-best \cite{TVGestel2004} & CCSS \cite{SDzeroski2004} & COOP \cite{NGPedrajas2005} & CNNE \cite{MIslam2003} & EPNet \cite{XinYao1997} & EENCL \cite{YLiu2000} \\ \hline
    Error Rate & 19.67 & 24.70 & 22.70 & 24.02 & 19.69 & \textbf{19.60} & 22.38 & 22.1  \\ \hline
  \end{tabular}%
  \end{center}
\end{table*}

The Wisconsin Breast Cancer dataset contains 699 samples, each of which has real-valued attributes and can be classified into two classes: 458 \textit{benign} and 241 \textit{malignant}. To test the performance of CROANN, all samples are divided into three parts by simple random sampling method: 349 training samples, 175 validation samples and 175 testing samples.

Results from CROANN and six other ANNs are listed in Table IV. CROANN performs superior to SGAANN, EPANN, PSOANN, GSOANN, and MGNN, and it has a similar performance with ESANN. As compared with GSOANN, CROANN can give a better standard derivation. There are also comparisons with recent published results listed in Table V. We also compare CROANN with other machine learning algorithms and the results are given in Table V. We can see that CROANN performs very well and shares the best performance with GANet-best.

\subsubsection{Pima Indian Diabetes Dataset}

The Pima Indian Diabetes dataset contains 768 samples, 500 of which are indicated with sign of diabetes and 268 are without such sign. There are eight real-valued attributes that can be used to determine whether a patient has the sign of diabetes or not. This dataset is known as a difficult problem for machine learning for its scarcity of samples and heavy noise pollution. This dataset is partitioned into 384 training samples, 192 validation samples and the 192 testing samples.

Table VI shows the comparison between CROANN and five other EA-based ANNs (MGNN is not included because there is no simulation data provided for this dataset in \cite{PPPalmes2005}). CROANN again outperforms the rest in terms of the testing set mean error. Results from other machine learning algorithms are tabulated in Table VII, which demonstrates that CROANN achieves a performance that is comparable with the best, i.e. CNNE, and greatly outperforms the others.

\section{Conclusion}

In this paper, we propose a novel EA-based ANNs called CROANN, to train ANNs, based on CRO. We have shown that CROANN can optimize the structure as well as the weights of ANNs simultaneously using stochastic processes. In CROANN, the structure and weights of a network is encapsulated in one solution, which is considered as a point in the solution space. In this way, CROANN searches the global minimum, which represents the network configuration providing the best performance. Since there are no restrictions on the evolution of the network structure and on the weight adaptation, CROANN does not suffer from the "structural hill-climbing" problem observed in the constructive and pruning approaches of ANN \cite{PPPalmes2005}. Simulation results show that CROANN can outperform other existing EA-based ANN algorithms. In the Iris dataset and the Pima Indian Diabetes dataset, CROANN provides the best testing error rate among all representative EA-based ANNs. Although CROANN is not the best in the test with the Wisconsin Breast Cancer dataset, the gap between the result generated by CROANN and the best one is indeed very small. In the comparisons with other sophisticated machine learning algorithms on the three classification problems, CROANN can always provide the best performance. To summarize, CRO is well suited to be incorporated in ANN to solve classification problems. In the future, we will conduct a systematic analysis of variance on the parameters and perform Student's t-test to show significance of the results. Moreover, we can further explore the ability of CROANN to solve some real-world classification problems, and other types of problems including function approximation, data processing, and robotics.

\section*{Acknowledgement}

This work is supported in part by the Strategic Research Theme of Information Technology of The University of Hong Kong. A.Y.S. Lam is also supported in part by the Croucher Foundation Research Fellowship.

\end{document}